\documentclass{article}

\PassOptionsToPackage{numbers, sort&compress}{natbib}

 \usepackage[preprint]{neurips_2026}

\workshoptitle{Foundation Models for the Brain and Body}

\usepackage[utf8]{inputenc} %
\usepackage[T1]{fontenc}    %
\usepackage{hyperref}       %
\usepackage{url}            %
\usepackage{booktabs}       %
\usepackage{amsfonts}       %
\usepackage{nicefrac}       %
\usepackage{microtype}      %
\usepackage{xcolor}         %
\usepackage{graphicx}       %
\usepackage{wrapfig}        %
\graphicspath{{figs/}}      %
\usepackage{flafter}        %

\title{Pretraining for Sample-Efficient Neural Interfaces}

\author{%
  Ben Tang\quad
  Zachary Spalding\quad
  Gregory B. Cogan\thanks{Corresponding author: \texttt{gregory.cogan@duke.edu}. Code: \url{https://github.com/bentang18/MAPA}} \\[0.4em]
  \mdseries Duke University, Durham, NC \\
  \mdseries\texttt{\{ben.tang,zac.spalding,gregory.cogan\}@duke.edu}
}

\begin{document}

\maketitle

\begin{abstract}
Brain-computer interfaces (BCIs) decode neural activity to restore lost function. Typically, training a high-performance neural decoder requires a large labeled dataset to be collected from every new subject. One way to reduce the labeled data cost is self-supervised pretraining, which learns general neural representations from unlabeled recordings that accumulate across subjects. However, for intracranial electroencephalography (iEEG) recordings, self-supervised learning has been challenging due to differences in contact placement and neuroanatomy between subjects. We propose MAPA, an otherwise vanilla masked autoencoder with two spatial encodings, an anatomical region embedding and a relative positional encoding, which together enable it to learn neural representations that transfer to unseen subjects and across various tasks. MAPA sets a new state of the art across all three regimes of the Neuroprobe benchmark without fine-tuning: within-session, cross-session, and cross-subject. In the cross-subject regime, a linear probe on MAPA's features needs only ${\sim}164$ labeled trials to reach the accuracy that takes 3,500 without pretraining. Our results show that self-supervised pretraining can scale across heterogeneous iEEG recordings and reduce the labeled data needed for accurate decoding in new subjects.
\end{abstract}

\section{Introduction}
\begin{figure}[t]
  \centering
  \includegraphics[trim=0 0 10.435bp 0, clip, width=\textwidth]{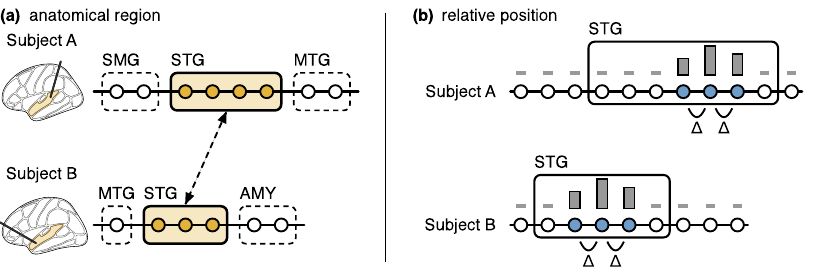}
  \caption{\textbf{Two spatial properties shared across subjects.} (a) Arrays with differing placement across subjects sample the same neuroanatomical regions. (b) The same pattern of activity falls at different positions along the array in each subject.}
  \label{fig:correspondence-priors}
\end{figure}

A brain-computer interface (BCI) can decode neural activity into movement, text, or speech, restoring function lost to neurological injury or disease \citep{leuthardt2004brain, willett2021highperformance, metzger2023highperformance}. However, current neural decoders require a large labeled dataset from every new subject to reach high accuracy on fine-grained tasks. Collecting labeled data takes considerable time and effort, which hinders the successful deployment of BCIs, especially for people living with stroke or advanced neuromuscular disease \citep{cumming2016prevalence, hamad2023prevalence}.

Recent work has proposed training models on data pooled across subjects to improve accuracy and reduce calibration cost. Two separate approaches have emerged in the literature: pooling labeled data with supervised learning, and pooling unlabeled data with self-supervised learning. Prior work in the supervised lineage \citep{mentzelopoulos2024neural, chen2025transformerbased, singh2025transfer, levin2026crossbrain, fogg2026generalizable, spalding2026shared} has shown improvements in decoding performance; however, scaling supervised models is still limited by the amount of labeled data available, which grows only with the hours a subject can spend on a task. On the other hand, self-supervised models learn directly from raw recordings that accumulate continuously, without needing an experimental paradigm or additional subject participation. Despite the advantage in data scale, pretrained models have so far shown only marginal gains over task-specific baselines \citep{banville2026neuralbench, kontras2026neuroatlas}.

Here, we explore self-supervised pretraining for intracranial EEG (iEEG), which records neural activity from electrode arrays placed on the brain surface in electrocorticography (ECoG) or inside brain tissue in stereo-electroencephalography (sEEG). Intracranial recordings sample field potentials at high fidelity and fine spatial resolution \citep{parvizi2018promises}, and accumulate throughout a clinical monitoring stay, producing large volumes of unlabeled data from many subjects \citep{peterson2022ajile12, carzaniga2026foundation, evanson2025minutes}. For these reasons, iEEG is a strong candidate for self-supervised pretraining, but \emph{differences in contact placement and neuroanatomy} between subjects have made pooling iEEG recordings difficult.

Nonetheless, two spatial properties are shared across subjects even when individual sampling differs. First, function is largely localized in the brain: neuroanatomical regions that support movement, language, and perception are conserved across people \citep{glasser2016multimodal}, so contacts sampling the same region in different subjects record similar activity during the same task (Figure~\ref{fig:correspondence-priors}a). Second, the same pattern of activity falls at different absolute positions along an array in each subject, but a pattern is defined by its relative position across contacts (Figure~\ref{fig:correspondence-priors}b). Crucially, the anatomical region of a contact and its relative position within an array are known for every subject, wherever the array is placed.

We design MAPA (Masked Autoencoder with Positional and Anatomical encodings) around these two spatial properties by applying an anatomical region embedding to each contact, and a relative positional encoding between contacts in an array. Prior work either did not encode spatial position \citep{wang2023brainbert}, or encoded it as a physical coordinate \citep{chau2025population}, as a learned embedding per channel \citep{mahato2025scalable}, or as a coordinate in a template brain \citep{han2026diver1}. BaRISTA \citep{oganesian2025barista} found that encoding position at the scale of regions improves downstream decoding. The relative positional encoding within an array is, to our knowledge, novel for iEEG models. The rest of MAPA is a vanilla vision transformer (ViT) \citep{dosovitskiy2021image} pretrained by masked autoencoding \citep{he2022masked}, held at its default configuration to isolate the effects of our spatial encodings.

We evaluate on Neuroprobe \citep{zahorodnii2026neuroprobe}, a benchmark for iEEG foundation models scored over 15 tasks spanning audio, language, and vision decoding during movie-watching. The benchmark defines three regimes: within-session, cross-session, and cross-subject. For pretraining we use 27.9 hours of Brain Treebank \citep{wang2024brain} recordings from 13 sessions of 7 subjects, matching the data used by most prior entries. To investigate how pretraining transfers to an unseen subject, we hold out two evaluation subjects from pretraining entirely.

Our main contributions are as follows:
\begin{itemize}
    \item \textbf{New state of the art on Neuroprobe.} MAPA ranks first in all three regimes, and in cross-session and cross-subject it is the only pretrained entry above the task-specific baselines.
    \item \textbf{Pretraining gain from the spatial encodings.} With neither spatial encoding, MAPA falls below our frontend baseline in all regimes.
    \item \textbf{Cross-subject decoding with 21\texttimes\ fewer labels.} A linear probe on MAPA's frozen features reaches the frontend baseline's full-data accuracy with ${\sim}164$ labeled trials rather than 3,500.
\end{itemize}

\section{Method}

\begin{figure}[ht]
  \centering
  \includegraphics[width=\textwidth]{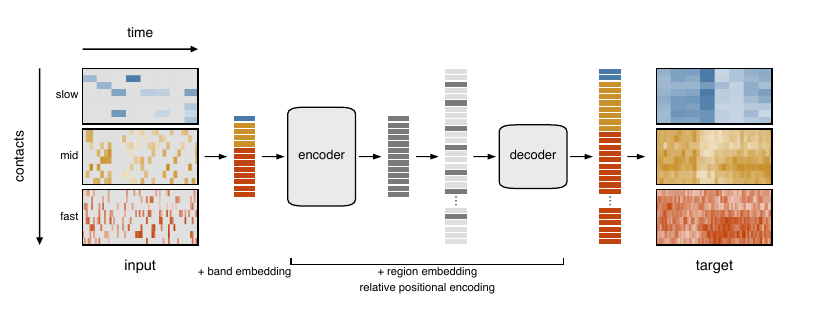}
  \caption{\textbf{MAPA architecture.} During pretraining, a large random subset of patches over electrode contacts, frequency bands, and time is masked out. The encoder operates on the visible patches only, and a small decoder then processes the full set of encoded patches and mask tokens to reconstruct the frequency bins of the removed patches. Our two spatial encodings enter both the encoder and the decoder, and the decoder is discarded after pretraining.}

  \label{fig:methods}
\end{figure}

MAPA is an extension of Masked Autoencoders (MAE) \citep{he2022masked}, a self-supervised learning method, to intracranial neural recordings (Figure~\ref{fig:methods}). The encoder, decoder, and objective follow the standard MAE design. We change only the patching and the two spatial encodings to fit iEEG data.

\setlength{\intextsep}{0pt}
\begin{wraptable}[12]{r}{0.54\textwidth}
\setlength{\abovecaptionskip}{0pt}
\setlength{\belowcaptionskip}{7pt}
\centering
\caption{\textbf{Input representation.} Each band contains the magnitude of a short-time Fourier transform at the listed window length. Long windows resolve low frequencies, and short windows resolve high frequencies. These frequency bins are the input to both the encoder and the frontend baseline.}
\label{tab:bands}
\begin{tabular}{lrrrr}
\toprule
Band & Window & Rate & Frequencies & Bins \\
\midrule
Slow & 500 ms   & 4 Hz  & 2--14 Hz   & 7 \\
Mid  & 125 ms   & 16 Hz & 16--56 Hz  & 6 \\
Fast & 62.5 ms  & 32 Hz & 64--160 Hz & 7 \\
\bottomrule
\end{tabular}
\end{wraptable}
\paragraph{Frontend.}
Given an array of $N$ contacts (an sEEG shaft or an ECoG grid), we reference the raw voltage of each contact to the mean over the array. We then take the magnitude of three short-time Fourier transforms, each at a different window length (Table~\ref{tab:bands}), to capture neural processes that differ in frequency and timescale \citep{buzsaki2004neuronal}. Finally, we normalize each frequency bin per contact by its median value and median absolute deviation over a session, and clip the normalized bins to a fixed range (Appendix~\ref{app:guards}).
\setlength{\intextsep}{12pt plus 2pt minus 2pt}

\paragraph{Patching.}
A patch is the frequency bins of one band, at one contact, at one time step of that band's rate. Over $T$ seconds, each contact contributes $4T$ Slow, $16T$ Mid, and $32T$ Fast patches. A per-band linear layer projects each patch to a token at the model width. We add a learned band embedding to each token.

\paragraph{Positional encoding.}
We apply two positional encodings. The first is a learned region embedding applied to each token, where regions are supplied by a standard brain atlas \citep{klein2012101} (Appendix~\ref{app:region}). The second is a rotary position embedding (RoPE) \citep{su2024roformer}, applied to queries and keys on the spatial axis of the array and on time. Position along the array is the integer contact number read from the clinical label, and RoPE encodes only the difference between two contact numbers.

\paragraph{Masking.}
We mask patches uniformly at random, with no structure over contacts, bands, or time. We remove 75\% of all patches, the ratio used in the original MAE paper \citep{he2022masked}, though the optimal ratio depends on the redundancy between patches \citep{feichtenhofer2022masked}, which is set by contact spacing and the overlap between successive windows (Table~\ref{tab:bands}). We hold our masking ratio constant across every run.

\paragraph{Autoencoding.}
Our encoder is a ViT-Small \citep{dosovitskiy2021image} (21.3M parameters), applied only on the visible set of patches. Our decoder is a second ViT at half the depth and width, applied on the union of the encoded patch set and a set of mask tokens. We apply the band embedding, region embedding, and relative positional encoding in the decoder as in the encoder. The decoder reconstructs the frequency bins of each removed patch through a per-band linear layer, and the loss is the mean squared error over removed patches. Following V-JEPA 2.1 \citep{murlabadia2026vjepa}, we supervise the encoder at intermediate layers as well as its output (Appendix~\ref{app:model}). The decoder is discarded after pretraining.

\section{Experiments}

\subsection{Setup}

Our core experiment keeps the frontend and the readout fixed, with and without the encoder. The \emph{frontend baseline} is frontend $\rightarrow$ readout, and MAPA is frontend $\rightarrow$ encoder $\rightarrow$ readout. The encoder is pretrained and then frozen for all evaluations, and the readout is a linear probe (Appendix~\ref{app:readout}).

We evaluate both models on Neuroprobe, which scores 15 binary tasks by the area under the receiver operating characteristic curve (AUROC). We report macro AUROC, the mean AUROC over the 15 tasks (Appendix~\ref{app:pertask}). Each trial is the window from 0 to 1\,s after a word onset. Evaluation covers 12 sessions from 6 subjects, 2 sessions per subject, with at most 3,500 trials per task (Appendix~\ref{app:data}).

Neuroprobe includes three evaluation regimes, and they differ in what the readout is fit on. Within-session fits on one contiguous half of a session and tests on the other half. Cross-session fits on one session of a subject and tests on the held-out session. Cross-subject fits on one anchor subject and tests on each of the remaining subjects.

\subsection{Results on Neuroprobe}

\setlength{\intextsep}{0pt}
\begin{wraptable}[13]{r}{0.51\textwidth}
\setlength{\abovecaptionskip}{0pt}
\setlength{\belowcaptionskip}{7pt}
\centering
  \caption{\textbf{Neuroprobe leaderboard.} Macro AUROC over 15 tasks. Only the best entry of each published method family is shown.}
  \label{tab:leaderboard}
  \begin{tabular}{@{}lccc@{}}
Method & \shortstack{Within\\session} & \shortstack{Cross\\session} & \shortstack{Cross\\subject} \\
\midrule
BrainBERT & 0.626 & 0.633 & 0.547 \\
CNN Lap+STFT & 0.669 & \underline{0.670} & \underline{0.578} \\
PopT & 0.670 & 0.663 & 0.575 \\
DIVER-1 frozen & \underline{0.678} & $\times$ & $\times$ \\
\midrule
Frontend baseline & 0.674 & 0.666 & 0.587 \\
\textbf{MAPA} & \textbf{0.695} & \textbf{0.691} & \textbf{0.608} \\
\end{tabular}

\end{wraptable}
MAPA ranks first in all three regimes (Table~\ref{tab:leaderboard}). Outside the within-session regime, MAPA is the only pretrained model above the best task-specific baseline (CNN Lap+STFT). MAPA's lead over published entries is largest in the cross-subject regime.

Our frontend baseline is strong on its own, placing above 12 of the 13 within-session entries, 11 of 12 cross-session, and all 10 cross-subject. MAPA improves on it in every regime. Our two entries share their frontend and readout, so the \emph{gain comes from the pretrained encoder alone}.

\setlength{\intextsep}{12pt plus 2pt minus 2pt}
\subsection{Ablation of the two spatial encodings}

\begin{figure}[!ht]
  \centering
  \includegraphics[width=\textwidth]{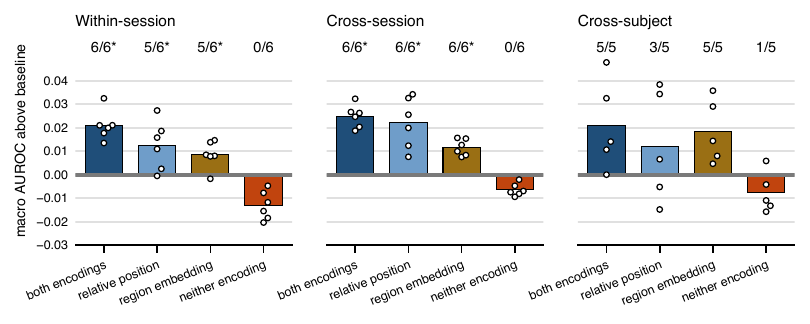}
  \caption{\textbf{Gain from spatial encodings.} Macro AUROC gain over the frontend baseline, one dot per subject, with the number of subjects improved above each bar. $*$ marks $q < 0.05$ against the baseline, one-sided paired permutation test over subjects (Appendix~\ref{app:stats}).}
  \label{fig:geometry}
\end{figure}

To test how the gain depends on the two spatial encodings, we pretrain three more models that remove each encoding alone and both together, and read out all four identically (Figure~\ref{fig:geometry}). Both encodings together improve over the baseline on every subject in all three regimes, and the improvement is significant within-session and cross-session (Appendix~\ref{app:stats}). Each encoding alone improves less than both encodings together. With neither, the pretrained model falls below the baseline.

\subsection{Generalization to held-out subjects}
\setlength{\intextsep}{0pt}
\begin{wrapfigure}[19]{r}{0.39\textwidth}
  \centering
  \includegraphics[width=0.9648\linewidth]{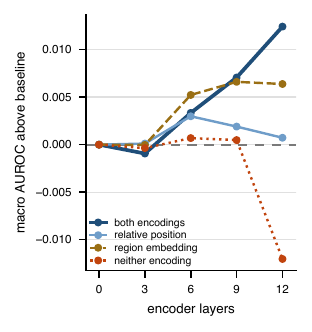}
  \caption{\textbf{Gain in held-out subjects.} Macro AUROC by encoder depth in the cross-subject regime, on the two subjects excluded from pretraining. Layer 0 is the frontend baseline.}
  \label{fig:zeropretrain}
\end{wrapfigure}
The strictest test of cross-subject transfer fits the readout on an anchor subject and evaluates on subjects held out from pretraining. Since neither the encoder nor the readout is fit on the held-out subjects, a gain requires learning \emph{neural representations that generalize across subjects}.

In the cross-subject regime, MAPA improves macro AUROC over the frontend baseline on both of our held-out subjects (Appendix~\ref{app:heldout}). With neither spatial encoding, the model falls below the baseline on both subjects (Figure~\ref{fig:zeropretrain}). The four models stay close through the first half of the encoder and separate in the deeper layers, where the gain keeps rising only with both encodings.
\setlength{\intextsep}{12pt plus 2pt minus 2pt}

\subsection{Sample efficiency}
\begin{figure}[!ht]
  \centering
  \includegraphics[width=\textwidth]{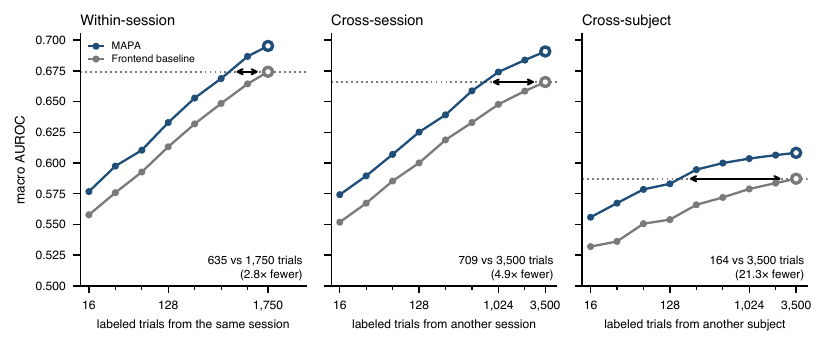}
  \caption{\textbf{Pretraining reduces the labels needed for a given accuracy.} Macro AUROC against the number of labeled trials on a log axis. The arrow marks the label saving against the frontend baseline fit on the full training set. MAPA is above the baseline at every label count we test, and the gain does not close as labels grow (Appendix~\ref{app:labels}).}
  \label{fig:sample_efficiency}
\end{figure}

To test whether pretraining reduces the labels needed to reach a given accuracy, we refit the readout on subsampled label counts (Appendix~\ref{app:labels}). In the cross-subject regime, MAPA reaches the frontend baseline's full-data accuracy with ${\sim}164$ labeled trials rather than 3,500, a 21\texttimes\ saving (Figure~\ref{fig:sample_efficiency}).

\section{Discussion}
Simple methods that scale are at the core of deep learning. However, two obstacles have prevented scaling models on intracranial neural recordings. First, labels are scarce, since every trial demands effort from an implanted subject. Second, pooling recordings is difficult, as contact placement and neuroanatomy differ between subjects. We address both obstacles by showing that a masked autoencoder, given an anatomical region embedding and a relative positional encoding, learns, directly from unlabeled recordings, neural representations that transfer across subjects.

Now is the time to scale. Intracranial recordings are accumulating en masse across institutions and experimental paradigms and can be pooled into one corpus through self-supervised learning and our two spatial encodings. Simple methods trained on large amounts of data have produced powerful general models in language, audio, and vision. We hope the same can be done for neural recordings.

\begin{ack}
This work used the Delta and DeltaAI systems at the National Center for Supercomputing Applications through allocation CIS261108 from the Advanced Cyberinfrastructure Coordination Ecosystem: Services \& Support (ACCESS) program \citep{boerner2023access}, which is supported by U.S. National Science Foundation grants \#2138259, \#2138286, \#2138307, \#2137603, and \#2138296. This research used both the DeltaAI advanced computing and data resource, which is supported by the National Science Foundation (award OAC 2320345) and the State of Illinois, and the Delta advanced computing and data resource, which is supported by the National Science Foundation (award OAC 2005572) and the State of Illinois. Delta and DeltaAI are joint efforts of the University of Illinois Urbana-Champaign and its National Center for Supercomputing Applications. Z.S. and G.B.C. were supported by NIH R01DC019498. G.B.C. was supported by NIH R01NS129703.

We thank Geeling Chau, Christopher Wang, Shivashriganesh P. Mahato, Saba Hashemi, and Andrii Zahorodnii for helpful discussions.

\end{ack}

\bibliographystyle{plainnat}
\bibliography{refs}

\clearpage
\appendix
\section*{Appendix}

Brain Treebank \citep{wang2024brain} is a dataset of sEEG recordings from subjects watching Hollywood movies. Neuroprobe \citep{zahorodnii2026neuroprobe} defines the tasks, trials, and splits over those recordings. We pretrain on the entire recording of each pretraining session. We evaluate on Neuroprobe Lite, the default form of the benchmark, which includes 12 sessions, at most 120 contacts per subject, and at most 3,500 trials per task. A session is one subject watching one movie. A shaft is one implanted sEEG electrode, the array the frontend references over, and a contact is one recording site along it.

\section{Data and splits}
\label{app:data}

\setlength{\intextsep}{0pt}
\begin{wraptable}[43]{r}{0.57\textwidth}
\setlength{\abovecaptionskip}{0pt}
\setlength{\belowcaptionskip}{7pt}
\centering
\caption{\textbf{Every Brain Treebank session.} Each row is one session, written as the subject number and the index of the session within that subject. Split marks the session as a pretraining session or a test session. Shafts, Contacts, and Regions count the shafts, contacts, and anatomical regions of the session that enter the model. Hours is the recording time of a pretraining session that enters pretraining. A subject keeps the same implant across sessions, so the same shaft appears in every session of the subject. The Total rows do not sum the columns above them: each counts every unique shaft, contact, and region once over the sessions of the split, and sums the hours.}
\label{tab:sessions}
\small
\begin{tabular*}{\linewidth}{@{\extracolsep{\fill}}llrrrr@{}}
\toprule
Session & Split & Shafts & Contacts & Regions & Hours \\
\midrule
1:0  & Pretrain & 13 & 128 & 16 & 1.91 \\
1:1  & Test     & 13 & 119 & 16 & ---  \\
1:2  & Test     & 13 & 119 & 16 & ---  \\
\midrule
2:0  & Test     & 14 & 119 & 14 & ---  \\
2:1  & Pretrain & 16 & 133 & 14 & 2.42 \\
2:2  & Pretrain & 16 & 132 & 14 & 2.66 \\
2:3  & Pretrain & 16 & 135 & 14 & 3.00 \\
2:4  & Test     & 14 & 119 & 14 & ---  \\
2:5  & Pretrain & 16 & 135 & 14 & 1.85 \\
2:6  & Pretrain & 16 & 129 & 14 & 3.52 \\
\midrule
3:0  & Test     & 12 & 100 & 14 & ---  \\
3:1  & Test     & 12 & 102 & 14 & ---  \\
3:2  & Pretrain & 12 & 103 & 14 & 4.06 \\
\midrule
4:0  & Test     & 11 & 120 & 22 & ---  \\
4:1  & Test     & 11 & 118 & 22 & ---  \\
4:2  & Pretrain & 15 & 179 & 28 & 1.31 \\
\midrule
5:0  & Not used & 12 & 140 & 18 & ---  \\
\midrule
6:0  & Pretrain & 12 & 159 & 16 & 0.81 \\
6:1  & Pretrain & 12 & 159 & 16 & 1.32 \\
6:4  & Pretrain & 12 & 160 & 16 & 1.60 \\
\midrule
7:0  & Test     & 8  & 118 & 20 & ---  \\
7:1  & Test     & 8  & 116 & 19 & ---  \\
\midrule
8:0  & Pretrain & 13 & 149 & 20 & 1.41 \\
\midrule
9:0  & Pretrain & 12 & 91  & 13 & 2.00 \\
\midrule
10:0 & Test     & 12 & 119 & 19 & ---  \\
10:1 & Test     & 12 & 120 & 19 & ---  \\
\midrule
Total & Pretrain & 93 & 945 & 50 & 27.87 \\
Total & Test     & 70 & 698 & 49 & ---   \\
\bottomrule
\end{tabular*}
\end{wraptable}
Pretraining uses 13 of the 26 sessions, 27.87 hours of recording on 93 shafts. Neuroprobe Lite evaluates 12 sessions. Table~\ref{tab:sessions} lists all 26. Not every implanted contact is usable. By default, Neuroprobe excludes the contacts Brain Treebank marks as corrupted, the trigger channels that carry the movie timing signal rather than neural activity, and the contacts missing from the Brain Treebank localization table. A contact missing from the table has no position and no region label. We apply the same exclusions before any other step. Guard 1 of Appendix~\ref{app:guards} then drops the contacts whose voltage fails a quality test, and Table~\ref{tab:sessions} counts what remains. Hours are counted before Guard 2, which removes 0.30 hours.

Pretraining uses every remaining contact of a session, and evaluation uses a subset: Neuroprobe Lite includes at most 120 contacts of a subject. For evaluation we restrict a session to its evaluation contacts before the frontend takes the mean over the array. No excluded contact enters an evaluation, not even through the array mean re-reference.

\paragraph{Corpus sampling.}
We draw a subject uniformly, then a shaft uniformly within the subject, then a clip uniformly within the shaft.

\paragraph{Pretraining rules.}
Neuroprobe permits pretraining on 14 full sessions and 5 partial ones. We use 13 of the 14, omitting session 5:0, whose subject Neuroprobe does not evaluate. The five partial sessions are the parts of subjects 7 and 10 that the evaluation does not cover, about 20 minutes per subject. We hold subjects 7 and 10 out of pretraining, so we use none of the five partial sessions, though competing entries were permitted all of them.
\setlength{\intextsep}{12pt plus 2pt minus 2pt}

\newpage
\section{Pretraining recipe}
\label{app:model}

\setlength{\intextsep}{0pt}
\begin{wraptable}[21]{r}{0.42\textwidth}
\setlength{\abovecaptionskip}{0pt}
\setlength{\belowcaptionskip}{7pt}
\centering
\caption{\textbf{Pretraining recipe.} Hyperparameters of the reported run.}
\label{tab:recipe}
\small
\begin{tabular*}{\linewidth}{@{}l@{\extracolsep{\fill}}r@{}}
\toprule
Parameter & Value \\
\midrule
Clip length (s) & 2.0 \\
Masked fraction & 0.75 \\
Steps & 55,000 \\
Warmup steps & 5,000 \\
Cooldown steps & 5,000 \\
Learning rate & 6e-3 \\
$\beta_1$, $\beta_2$ & 0.9, 0.95 \\
Weight decay & 0.04 \\
Gradient clip & 3.0 \\
Decoder layers & 6 \\
Decoder embedding size & 192 \\
Encoder intermediate layers & 3, 6, 9, 12 \\
Subject sampling & uniform \\
Contacts per step (global) & 66,000 \\
Seed & 33 \\
\bottomrule
\end{tabular*}
\end{wraptable}

Table~\ref{tab:recipe} gives the hyperparameters. Pretraining runs for 55,000 steps. We train with AdamW \citep{loshchilov2019decoupled}. The learning rate warms up linearly over 5,000 steps, holds at 6e-3 for 45,000, and falls linearly to zero over the final 5,000.

We chose the number of training steps on pretraining sessions, not on test sessions. During the constant phase we fit the readout on four tasks in seven pretraining sessions, none of which Neuroprobe evaluates, and we start the cooldown once the readout stops improving. The cooldown step was set on an earlier run and the reported run keeps it. We report the checkpoint at the end of the run, with no selection by benchmark score.

A batch holds a fixed number of contacts rather than of clips. The global batch is 66,000 contacts per step. A contact only attends to other contacts on its shaft in the encoder and decoder. Model runs take 32 hours on two NVIDIA H100 GPUs.

Following V-JEPA 2.1 \citep{murlabadia2026vjepa}, we supervise the encoder at layers 3, 6, 9, and 12, concatenate the four outputs to $4d$ for $d = 384$, and project to the decoder width. Deep supervision changes the decoder input and not the loss.
\setlength{\intextsep}{12pt plus 2pt minus 2pt}

\paragraph{Anatomical region embedding.}
\label{app:region}
The anatomical region embedding table has 74 labels: the 62 cortical regions of the Desikan-Killiany-Tourville (DKT) atlas \citep{klein2012101} and 12 subcortical structures. A contact takes the label of the region it lies in, read from the table Brain Treebank ships for each subject. A contact whose label is not one of the 74 takes a reserved 75th row. The embedding is indexed by the label alone, never by the subject, session, or contact number, so a new subject reuses the rows of the regions its contacts fall in and adds no parameters. The region embedding of a contact is added to every token of that contact. The ablation of Figure~\ref{fig:geometry} removes the table. The table is initialized near zero, so at the start of pretraining the embedding contributes nothing and every row is learned.

Pretraining covers 50 of the 74 regions and the evaluation contacts fall in 49. Four regions, all in the right hemisphere, appear only in evaluation, so their rows are never trained and stay at their near-zero initialization.

\section{Readout protocol}
\label{app:readout}

\paragraph{Trials and tasks.}
A trial is the recording from 0 to 1\,s after a word onset. Neuroprobe defines 15 binary tasks, each a label of the word or of the movie at that onset. A session contributes at most 3,500 trials per task. A task is scored by AUROC. Trials, labels, splits, and evaluation contacts come from Neuroprobe unaltered. The published values in Table~\ref{tab:leaderboard} are from the Neuroprobe leaderboard, and our two entries follow the protocol below.

\paragraph{Readout input.}
We run the frozen encoder once over the trials of a session and cache the output of its last layer. The encoder runs on the shafts of a subject one at a time, and the readout gathers the outputs at all shafts. The readout input for one trial is the encoder output flattened over contacts, bands, and time. Within-session and cross-session keep the contacts separate. In the cross-subject regime no contact is shared between subjects, so we average the features of the contacts within each region and fit the readout on the regions shared by the anchor and test subjects, 3 to 7 per pair. The frontend baseline reads the encoder input, the bands of Table~\ref{tab:bands}, in place of the encoder output. Every step that follows is the same for both.

\paragraph{Ridge fit.}
We standardize each input dimension by the mean and standard deviation of the training trials. In the cross-subject regime the training trials come from the anchor, so the anchor's statistics are applied to the test subject. We then fit a ridge regression on the binary label in closed form.

\paragraph{Regularization.}
The regularization strength is the only quantity we select in our readout. We select it from a grid of 25 values spaced logarithmically from $10^{-4}$ to $10^{4}$, scaled by the mean squared norm of the training features. We use the value with the highest AUROC on the validation trials. A tie keeps the smallest value. The frontend baseline is fit by the same procedure over the same grid, so neither model is tuned harder than the other.

\paragraph{Regimes.}
Within-session fits on one contiguous half of a session and tests on the other half, in both directions, and the AUROC of a session is the mean of the two folds. Cross-session fits on one session of a subject and tests on the held-out session, in both directions. Cross-subject fits on session 2:4 of the anchor subject and tests on both evaluation sessions of each of the other five subjects, 10 test sessions in all. In every regime the validation and test trials are the first and second half of the test session, as Neuroprobe defines them. For each task we average the AUROC over the test sessions, 12 within-session and cross-session and 10 cross-subject. We report macro AUROC, the mean over the 15 tasks.

\section{Statistical methods}
\label{app:stats}

\paragraph{Subject-level tests.}
The claim is that the gain holds across subjects, so every test is at the subject level. The macro AUROC of a subject's two sessions is averaged into one value, so each subject contributes one gain. Within-session and cross-session have 6 subjects and cross-subject has 5, since subject 2 is the anchor.

\paragraph{Permutation test.}
We run an exact paired permutation test on the subject means. The difference for a subject is the macro AUROC of a model minus the macro AUROC of the frontend baseline, the gain from adding the pretrained encoder. The statistic is the mean difference over subjects. A permutation flips the sign of one subject's difference, and we enumerate all $2^n$ assignments. The test is one-sided, in the direction that a model exceeds the frontend baseline. An exact paired test on $n$ subjects cannot return a $p$ below $1/2^n$: 0.0156 at $n = 6$ and 0.0312 at $n = 5$.

\paragraph{Multiple comparisons.}
Within each regime, Figure~\ref{fig:geometry} compares four models against the frontend baseline, so we correct over those four with the Benjamini-Hochberg procedure \citep{benjamini1995controlling}. Table~\ref{tab:perm} gives every model. In the cross-subject regime no model clears $q < 0.05$, because the smallest $p$ an exact test on five subjects can return is 0.0312, and the correction over four models raises it to 0.0625.

\begin{table}[!ht]
\centering
\caption{\textbf{Every model against the frontend baseline.} Each model is one pretraining run, named for the spatial encodings it keeps. Subjects improved is the number of subjects whose macro AUROC exceeds the frontend baseline, out of the subjects in the regime. Gain is the mean over subjects of the difference in macro AUROC. $p$ is the one-sided exact paired permutation test and $q$ is the same $p$ after Benjamini-Hochberg correction within the regime. Bold marks $q < 0.05$.}
\label{tab:perm}
\small
\begin{tabular}{@{}llrrrr@{}}
\toprule
Regime & Model & Subjects improved & Gain & $p$ & $q$ \\
\midrule
Within-session & Both encodings    & 6/6 & $+0.0210$ & 0.0156 & \textbf{0.0417} \\
Within-session & Relative position & 5/6 & $+0.0125$ & 0.0312 & \textbf{0.0417} \\
Within-session & Region embedding  & 5/6 & $+0.0085$ & 0.0312 & \textbf{0.0417} \\
Within-session & Neither encoding  & 0/6 & $-0.0130$ & 1.0000 & 1.0000 \\
\midrule
Cross-session  & Both encodings    & 6/6 & $+0.0249$ & 0.0156 & \textbf{0.0208} \\
Cross-session  & Relative position & 6/6 & $+0.0221$ & 0.0156 & \textbf{0.0208} \\
Cross-session  & Region embedding  & 6/6 & $+0.0116$ & 0.0156 & \textbf{0.0208} \\
Cross-session  & Neither encoding  & 0/6 & $-0.0064$ & 1.0000 & 1.0000 \\
\midrule
Cross-subject  & Both encodings    & 5/5 & $+0.0211$ & 0.0312 & 0.0625 \\
Cross-subject  & Relative position & 3/5 & $+0.0119$ & 0.1875 & 0.2500 \\
Cross-subject  & Region embedding  & 5/5 & $+0.0184$ & 0.0312 & 0.0625 \\
Cross-subject  & Neither encoding  & 1/5 & $-0.0076$ & 0.9375 & 0.9375 \\
\bottomrule
\end{tabular}
\end{table}

\section{Per-task results}
\label{app:pertask}

Every macro AUROC in the main text is the mean over the 15 Neuroprobe tasks, and no task is dropped. Table~\ref{tab:pertask} gives the breakdown for the frontend baseline and for MAPA in all three regimes. MAPA exceeds the baseline on 41 of the 45 pairs of a task and a regime. The four pairs where the baseline is higher are frame brightness within-session and cross-subject, where both models are near chance, and pitch and global flow cross-subject, where the two models are within 0.004 of each other.

\begin{table}[!ht]
\centering
\caption{\textbf{The 15 Neuroprobe tasks.} AUROC for the frontend baseline and for MAPA, in each regime. Bold is the larger of a pair. Tasks appear in the order the benchmark lists them.}
\label{tab:pertask}
\small
\begin{tabular}{@{}lcccccc@{}}
\toprule
& \multicolumn{2}{c}{Within-session} & \multicolumn{2}{c}{Cross-session} & \multicolumn{2}{c}{Cross-subject} \\
\cmidrule(lr){2-3}\cmidrule(lr){4-5}\cmidrule(lr){6-7}
Task & Frontend & MAPA & Frontend & MAPA & Frontend & MAPA \\
\midrule
onset & 0.911 & \textbf{0.933} & 0.913 & \textbf{0.941} & 0.745 & \textbf{0.815} \\
speech & 0.891 & \textbf{0.917} & 0.909 & \textbf{0.932} & 0.719 & \textbf{0.779} \\
volume & 0.777 & \textbf{0.799} & 0.748 & \textbf{0.774} & 0.638 & \textbf{0.645} \\
delta volume & 0.778 & \textbf{0.800} & 0.768 & \textbf{0.787} & 0.674 & \textbf{0.705} \\
pitch & 0.597 & \textbf{0.628} & 0.590 & \textbf{0.633} & \textbf{0.540} & 0.537 \\
word index & 0.748 & \textbf{0.782} & 0.707 & \textbf{0.741} & 0.611 & \textbf{0.661} \\
word gap & 0.625 & \textbf{0.640} & 0.584 & \textbf{0.600} & 0.532 & \textbf{0.554} \\
gpt2 surprisal & 0.615 & \textbf{0.630} & 0.608 & \textbf{0.626} & 0.560 & \textbf{0.579} \\
word head pos & 0.610 & \textbf{0.625} & 0.606 & \textbf{0.628} & 0.563 & \textbf{0.593} \\
word part of speech & 0.601 & \textbf{0.628} & 0.608 & \textbf{0.643} & 0.537 & \textbf{0.545} \\
word length & 0.616 & \textbf{0.643} & 0.609 & \textbf{0.639} & 0.528 & \textbf{0.562} \\
global flow & 0.648 & \textbf{0.674} & 0.647 & \textbf{0.675} & \textbf{0.571} & 0.567 \\
local flow & 0.643 & \textbf{0.675} & 0.620 & \textbf{0.648} & 0.570 & \textbf{0.572} \\
frame brightness & \textbf{0.512} & 0.505 & 0.524 & \textbf{0.533} & \textbf{0.517} & 0.495 \\
face num & 0.545 & \textbf{0.552} & 0.550 & \textbf{0.563} & 0.503 & \textbf{0.515} \\
\midrule
Macro over 15 tasks & 0.674 & \textbf{0.695} & 0.666 & \textbf{0.691} & 0.587 & \textbf{0.608} \\
\bottomrule
\end{tabular}

\end{table}

\section{Held-out subjects}
\label{app:heldout}

Subjects 7 and 10 contribute no pretraining data, and in the cross-subject regime their readout is fit on the anchor subject, so their recordings enter neither pretraining nor the fit of the readout. As in every regime, the regularization strength is selected on the validation half of the test session (Appendix~\ref{app:readout}). Table~\ref{tab:heldout} gives the macro AUROC of the frontend baseline and of MAPA on their four sessions. MAPA improves the macro AUROC of every session, and therefore of both subjects. An exact paired test on two subjects cannot return a $p$ below $1/2^2 = 0.25$ (Appendix~\ref{app:stats}), so we report no significance test for the held-out subjects.

\begin{table}[!ht]
\centering
\caption{\textbf{The two subjects held out of pretraining.} Macro AUROC over the 15 tasks in the cross-subject regime. Sessions are named subject:session, and the last row is the mean over the four.}
\label{tab:heldout}
\small
\begin{tabular}{@{}lccc@{}}
\toprule
Session & Frontend & MAPA & Difference \\
\midrule
7:0 & 0.5234 & 0.5385 & $+0.0151$ \\
7:1 & 0.5326 & 0.5458 & $+0.0132$ \\
10:0 & 0.5754 & 0.5825 & $+0.0071$ \\
10:1 & 0.5994 & 0.6137 & $+0.0143$ \\
\midrule
Mean & 0.5577 & 0.5701 & $+0.0124$ \\
\bottomrule
\end{tabular}
\end{table}

\section{Label saving}
\label{app:labels}

\paragraph{Crossing.}
We ask how many labeled trials MAPA needs to match the frontend baseline fit on every labeled trial. We cut the training trials down to $m$, refit the readout on those $m$, and score on the full test trials, which gives macro AUROC as a function of $m$. The crossing is the smallest $m$ at which MAPA reaches the macro AUROC of the frontend baseline fit on every labeled trial, and the saving is the full count divided by the crossing. The full count is 1,750 trials within-session and 3,500 cross-session and cross-subject. In the cross-subject regime the crossing is 164 trials, the 21\texttimes\ saving the main text reports.

\paragraph{Subsampling.}
$m$ counts the training trials alone. The test trials are never subsampled, and the regularization strength is selected on the full validation trials. $m$ takes the values 16, 32, 64, 128, 256, 512, and 1,024, one more point at 2,048 cross-session and cross-subject, where the training set is large enough, and one point with no subsampling. At each $m$ we draw several random subsets of the training trials and average the resulting AUROCs: 5 subsets at the four smallest $m$, 3 at the larger $m$, and a single fit at the full count, where there is nothing to subsample. A subsample keeps the class balance of the training trials and is drawn once for each task, so both models are fit on the same trials. The crossing is interpolated in $\log_2 m$ between the two label counts that bracket it. If MAPA never reaches the frontend baseline's full-data macro AUROC, there is no crossing.

\paragraph{Confidence interval.}
The interval on the saving is a percentile bootstrap confidence interval with the subject as the resampling unit. A saving computed on five or six subjects is uncertain, and the bootstrap estimates how far the saving moves when the subjects are redrawn. We draw 20,000 resamples of the subjects with replacement, and both sessions of a subject enter a resample together. On each resample we recompute the macro AUROC at every $m$, the frontend baseline's full-data macro AUROC, and the saving. In 15 of the 20,000 cross-subject resamples there is no crossing. The interval is the 2.5th and 97.5th percentiles of the savings over all 20,000 resamples, with a resample that has no crossing ranked below every saving. The main text reports 21\texttimes\ as a point estimate with its interval.

\paragraph{Gain at every label count.}
We test 8 label counts within-session and 9 in cross-session and cross-subject. MAPA is above the frontend baseline at all 26 label counts we test across regimes, by 0.0178 to 0.0312 macro AUROC.

\paragraph{Per-subject saving.}
Table~\ref{tab:reach} gives the saving for each subject, computed the same way on the test sessions of that subject. The pooled row is computed on all test sessions together and is not the mean of the rows above it. In the cross-subject regime the saving ranges from 15.9 to 89.8. Subject 4 never reaches the frontend baseline at any label count we test, and it stays in the pooled saving and in the bootstrap.

\begin{table}[!ht]
\centering
\caption{\textbf{Label saving by subject.} The factor by which MAPA cuts the labeled trials needed to match the frontend baseline fit on every training trial. The pooled row is computed on all test sessions together and is not the mean of the rows above it. Subject 2 supplies the cross-subject training trials and is never a cross-subject test subject.}
\label{tab:reach}
\small
\begin{tabular}{@{}lrrr@{}}
\toprule
Subject & Within-session & Cross-session & Cross-subject \\
\midrule
1  & 4.9 & 12.3 & 28.7 \\
2  & 3.1 & 7.3  & $\times$ \\
3  & 2.5 & 4.1  & 89.8 \\
4  & 2.0 & 4.1  & never reaches \\
7  & 2.9 & 3.8  & 20.7 \\
10 & 2.3 & 3.1  & 15.9 \\
\midrule
Pooled & 2.8 & 4.9 & 21.3 \\
95\% interval & 2.3 to 3.5 & 3.7 to 7.3 & 8.4 to 64.0 \\
\bottomrule
\end{tabular}
\end{table}

\section{Data cleaning}
\label{app:guards}

We clean the data in three steps before any of it enters pretraining, and call each step a guard. Guard 1 removes contacts, Guard 2 removes time windows from pretraining, and Guard 3 bounds every value that enters the encoder. Guards 1 and 3 also apply to the evaluation sessions. Guard 2 applies to pretraining alone. No guard uses any task label, so the same rules apply to any recording. Every model trains on the same cleaned data. Table~\ref{tab:guards} lists every rule and the share of pretraining data it removes.

\paragraph{Preprocessing.}
Every recording is resampled to 2048 Hz. Each contact is notch filtered at 60 Hz and its harmonics and high-pass filtered at 0.5 Hz. Guard 1 runs on this voltage. A removed contact never enters the array mean.

\paragraph{Guard 1.}
Guard 1 runs before the frontend, on the voltage of every contact. Guard 1 has three rules. Spike removes a contact with sharp transients. Noisy removes a contact whose spread is far above the other contacts of the session. Dead removes a contact whose spread is far below them. Table~\ref{tab:guards} gives the thresholds. In the Noisy rule the denominator of $z$ is floored at 0.15 of the session median spread, so small differences within a uniform session do not fire the rule. A contact that fails any rule is removed for the whole session. Guard 1 removes 37 of the 1,829 pretraining contacts and 25 of the 1,414 evaluation contacts, counting a contact once per session. Guard 1 runs on each session separately, so a contact can fail in one session of a subject and pass in another.

\paragraph{Guard 2.}
Guard 2 runs on pretraining data only, after Guard 1 and the frontend and before Guard 3, on the three normalized bands (Table~\ref{tab:bands}). It cuts each session into 1-second windows and removes a window in which many contacts show an artifact at once, or one contact shows an extreme value. Thresholds are set per session and per band. In band $b$, we take the largest $|z|$ of each contact in each window, and $q_b$ is the 99th percentile of these values over all contacts and windows of the session. A rule that counts contacts fires when at least $n = \max(3, \lceil 0.05C \rceil)$ of the $C$ contacts of the session meet its condition in the same window. Guard 2 removes 1.08\% of 100,334 windows, which is 0.30 hours of the 27.87. It never touches an evaluation trial.

\paragraph{Guard 3.}
Guard 3 clips each normalized value as it enters the model, at $|z| = 15$ in the Slow and Mid bands and $|z| = 20$ in the Fast band, in pretraining and evaluation alike. It touches fewer than 0.03\% of the values in any band.

\begin{table}[!ht]
\centering
\caption{\textbf{The three guards.} A guard acts when any one of its rules fires. Rate is the share of pretraining data affected: of contacts for Guard 1, of windows for Guard 2, and of single values for Guard 3. The row of each guard gives the total share it removes. Spread is 1.4826 times the median absolute deviation, and $z$ is the distance from the median divided by the spread. Guard 1 runs on voltage, and $z$ is taken within a contact over time, except where a rule says over the contacts of the session. Guards 2 and 3 run on the normalized bands, where every value is already a $z$. The three rules of Guard 1 never fire on the same contact, so their rates sum to the share Guard 1 removes. A window can fail several rules of Guard 2, so those rates sum to more than the share Guard 2 removes. Each rule of Guard 3 acts on the values of one band, so the share Guard 3 removes is the mean of the three rates weighted by the number of values in each band.}
\label{tab:guards}
\small
\begin{tabular}{@{}l p{0.56\textwidth} r@{}}
\toprule
Rule & Fires when & Rate (\%) \\
\midrule
\multicolumn{2}{@{}l}{\textbf{Guard 1}} & \textbf{2.02} \\
\quad Spike & $|z|$ of the voltage slope exceeds 100 in over 1\% of the contact's 5-second clips & 1.37 \\
\quad Noisy & Spread, as a $z$ over the contacts of the session, exceeds 8 & 0.05 \\
\quad Dead  & Spread is below 0.35 of the median spread of the session & 0.60 \\
\midrule
\multicolumn{2}{@{}l}{\textbf{Guard 2}} & \textbf{1.08} \\
\quad Common mode  & $|z|$ exceeds $4q_b$ in $n$ contacts at once & 0.30 \\
\quad Catastrophic & One contact exceeds $8q_b$ in Slow or Mid, or $6q_b$ in Fast & 0.99 \\
\quad Absolute     & One contact exceeds $|z| = 200$ & 0.21 \\
\quad Dropout      & Slow band standard deviation is below 0.05 in $n$ contacts at once & 0.05 \\
\midrule
\multicolumn{2}{@{}l}{\textbf{Guard 3}} & \textbf{0.0152} \\
\quad Slow & $|z|$ exceeds 15 & 0.0103 \\
\quad Mid  & $|z|$ exceeds 15 & 0.0085 \\
\quad Fast & $|z|$ exceeds 20 & 0.0258 \\
\bottomrule
\end{tabular}
\end{table}

\end{document}